\documentclass{article}
\pdfoutput=1

\usepackage{graphicx} 
\usepackage{subfigure} 

\usepackage{amsmath}
\usepackage{algorithm}
\usepackage{algorithmic}

\usepackage{hyperref}

\newcommand{\fig}[1]{Figure~\ref{#1}}

\newcommand{\secn}[1]{Section~\ref{#1}}

\begin{document} 

\title{Semisupervised Classifier Evaluation\\ and Recalibration}
\author{Peter Welinder\footnote{
California Institute of Technology (\texttt{peter@welinder.se})}, 
Max Welling\footnote{
University of California, Irvine (\texttt{welling@ics.uci.edu })},
and Pietro Perona\footnote{
California Institute of Technology (\texttt{perona@caltech.edu})}}
\date{October 7, 2012}
\maketitle

\begin{abstract} 
How many labeled examples are needed to estimate a classifier's performance on a new dataset? We study the case where data is plentiful, but labels are expensive.  We show that by making a few reasonable assumptions on the structure of the data, it is possible to estimate performance curves, with confidence bounds, using a small number of ground truth labels. Our approach, which we call Semisupervised Performance Evaluation (SPE), is based on a generative model for the classifier's confidence scores. In addition to estimating the performance of classifiers on new datasets, SPE can be used to recalibrate a classifier by re-estimating the class-conditional confidence distributions.
\end{abstract} 

\section{Introduction}

Consider a biologist who downloads software for classifying the behavior of fruit flies. The classifier was laboriously trained by a different research group who labeled thousands of training examples to achieve satisfactory performance on a validation set collected in some particular setting (see e.g.~\cite{dankert09}). The biologist would be ill-advised if she trusted the published performance figures; maybe small lighting changes in her experimental setting have changed the statistics of the data and rendered the classifier useless. However, if the biologist has to review all the labels assigned by the classifier to her dataset, just to be sure the classifier is performing up to expectation, then what is the point of obtaining a trained classifier in the first place? Is it possible at all to obtain a reliable evaluation of a classifier when unlabeled data is plentiful, but when the user is willing to provide only a small number of labeled examples?

We propose a method for achieving minimally supervised evaluation of classifiers, requiring as few as 10 labels to accurately estimate classifier performance. Our method is based on a generative Bayesian model for the confidence scores produced by the classifier, borrowing from the literature on semisupervised learning~\cite{NigamEtal00,Seeger02,Zhu08}. We show how to use the model to re-calibrate classifiers to new datasets by choosing thresholds to satisfy performance constraints with high likelihood. An additional contribution is a fast approximate inference method for doing inference in our model.

\begin{figure*}[t]
\begin{center}
\includegraphics[width=\textwidth]{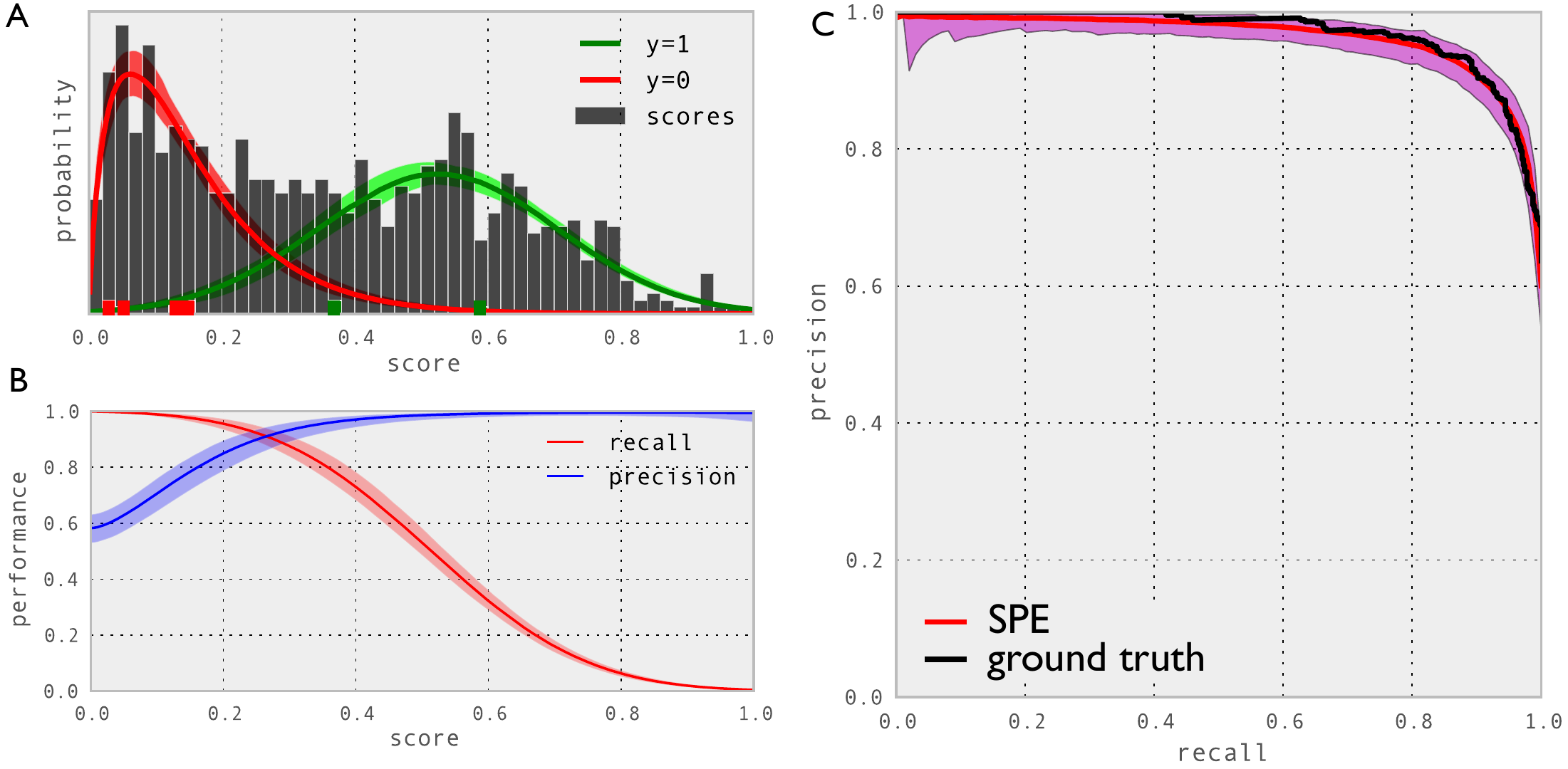}
\caption{Estimating detector performance with all but 10 labels unknown. {\bf A}: Histogram of classifier scores $s_i$ obtained by running the ``ChnFtrs'' detector \cite{DollarEtal11} on the INRIA dataset \cite{DalalTriggs05}. The red and green curves show the Gamma-Normal mixture model fitting the histogrammed scores with highest likelihood. The scores are all unlabeled, apart from 10, selected at random, which have labels. The shaded bands indicate the 90\% probability bands around the model. The red and green bars show the labels of the 10 randomly sampled labels (by chance, the scores for some of the samples are close to each other, thus only 6 bars are shown; the height of the bars has no meaning). {\bf B}: Precision and recall curves computed from the mixture model in A. {\bf C}:  In black, precision-recall curve computed after all items have been labeled. In red, precision-recall curve estimated using SPE from only 10 labeled examples (with 90\% confidence interval shown as the magenta band). See \secn{s:model} for a discussion.}
\label{f:toy-example}
\end{center}
\end{figure*}

\begin{figure*}[t]
\begin{center}
\includegraphics[width=\textwidth]{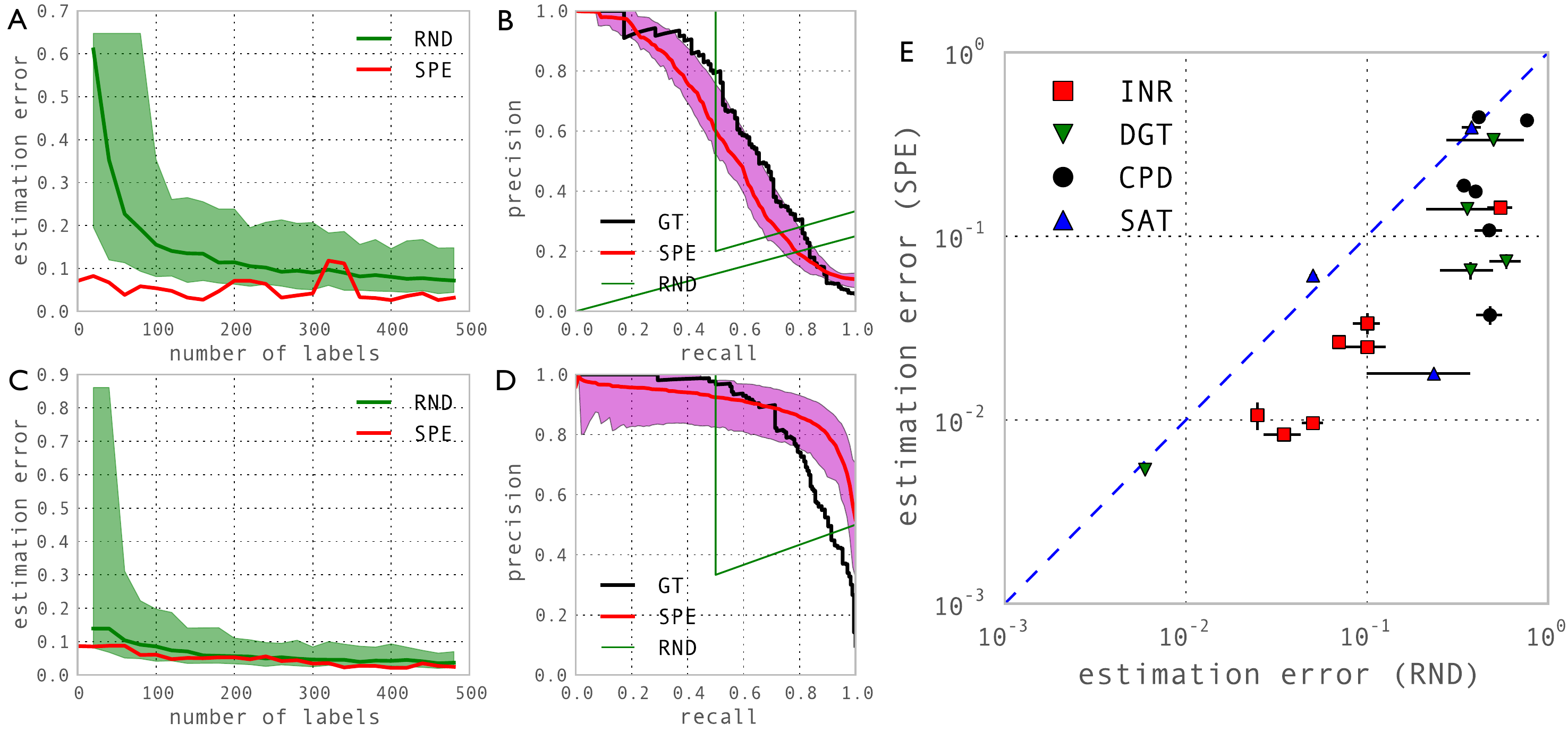}
\caption{Applying SPE to different datasets. \textbf{A}: Estimation error, as measured by the area between the true and predicted precision-recall curves, versus the number of labels sampled, for the ChnFtrs detector on the CPD dataset. The red curve is SPE and the green curve shows the median error of the naive method (RND). The green band show the 90\% quantiles of the naive method. \textbf{B}: The performance curve estimated using SPE (red) with 90\% confidence intervals (magenta) with 20 known labels. The ground truth performance with all label known is shown as a black curve (GT), and the performance curve computed on 20 labels using the naive method from 5 random samples is shown in green (RND). Notice that the curves (in green) obtained from different samples vary a lot (although most predict perfect performance). \textbf{C}--\textbf{D}: same as A--B, but for the logres8 classifier on the DGT dataset (hand-picked as an example where SPE does not work well). \textbf{E}: Comparison of estimation error (area between curves) of SPE and naive method for 20 known labels and different datasets. The appearance of the markers denote the dataset (each dataset has multiple classifiers), and the lines indicate the standard error averaged over 10 trials. SPE almost always perform significantly better than the naive method.}
\label{f:q-scheme-qual}
\end{center}
\end{figure*}

\section{Modeling the classifier score} \label{s:model}
Let us start with a set of $N$ data items, $(x_i, y_i) \in \mathcal{R}^D \times \{0,1\}$, drawn from some unknown distribution $p(x, y)$ and indexed by $i \in \{1, \ldots, N\}$. Suppose that a classifier, $\bar{h}(x_i; \tau) = [h(x_i) > \tau]$, where $\tau$ is some scalar threshold, has been used to classify all data items into two classes, $\hat{y}_i \in \{0,1\}$. While the ``ground truth'' labels $y_i$ are assumed to be unknown, initially, we do have access to all the ``scores,'' $s_i = h(x_i)$, computed by the classifier. From this point onwards, we forget about the data vectors $x_i$ and concentrate solely on the scores and labels, $(s_i, y_i) \in \mathcal{R} \times \{0,1\}$. 

The key assumption in this paper is that the list of scores $S=(s_1, \ldots, s_N)$ and the unknown labels $Y=(y_1,\ldots,y_N)$ can be modeled by a two-component mixture model $p(S,Y\mid\theta)$, parameterized by $\theta$, where the class-conditionals are standard parametric distributions. We show in \secn{s:fit-dists} that this is a reasonable assumption for many datasets.

Suppose that we can ask an expert (the ``oracle'') to provide the true label $y_i$ for any data item. This is an expensive operation and our goal is to ask the oracle for as few labels as possible. The set of items that have been labeled by the oracle at time $t$ is denoted by $\mathcal{L}_t$ and its complement, the set of items for which the ground truth is unknown, is denoted $\mathcal{U}_t$. This setting is similar to semisupervised learning \cite{Seeger02,Zhu08}. By estimating $p(S,Y\mid\theta)$, we will improve our estimate of the performance of $\bar{h}$ when $|\mathcal{L}_t|\ll N$. 

Consider first the fully supervised case, i.e. where all labels $y_i$ are known. Let the scores $s_i$ be i.i.d. according to the two mixture model. If the all labels are known, and we assume independent observations, the likelihood of the data is given by,
\begin{equation}
	p(S, Y \mid \theta) 
		=  \prod_{i:y_i = 0} (1-\pi) p_0(s_i \mid \theta_0)
		\prod_{i:y_i = 1} \pi p_1(s_i \mid \theta_1),
\label{e:obsv-likelihood}
\end{equation}
where $\theta = \{\pi, \theta_0, \theta_1\}$, and $\pi\in[0,1]$ is the mixture weight, i.e. $p(y_i=1)=\pi$. The component densities $p_0$ and $p_1$ could be modeled parametrically by Normal distributions, Gamma distributions, or some other probability distributions appropriate for the given classifier (see \secn{s:fit-dists} for a discussion about which class conditional distributions to choose). This approach of applying a generative model to score distributions, when all labels are known, has been used in the past to obtain error estimates on classifier performance \cite{HellmichEtal99,ErkanliEtal06,GuEtal08}, and for classifier calibration \cite{Bennett02}. However, previous approaches require that the all items used to estimate the performance have been labeled.

We suggest that it may be possible to estimate classifier performance even when only a fraction of the ground truth labels are known. In this case, the labels for the unlabeled items $i \in \mathcal{U}_t$ can be marginalized out,
\begin{align}
	p(S, Y_{t} \mid \theta) 
		&= \prod_{i \in \mathcal{U}_t} 
			\left((1-\pi) p_0(s_i \mid \theta_0)
				+ \pi p_1(s_i \mid \theta_1) \right) \\
		&\phantom{=}\times
		 \prod_{i \in \mathcal{L}_t} 
			\pi^{y_i}(1-\pi)^{1-y_i} p_{y_i}(s_i \mid \theta_{y_i}),
\label{e:likelihood}
\end{align}
where $Y_{t} = \{y_i \mid i\in\mathcal{L}_t\}$. This allows the model to make use of the scores of unlabeled items in addition to the labeled items, which enables accurate performance estimates with only a handful of labels. Once we have the likelihood, we can take a Bayesian approach to estimate the parameters $\theta$. Starting from a prior on the parameters, $p(\theta)$, we can obtain a posterior $p(\theta \mid S,Y_t)$ by using Bayes' rule,
\begin{equation}
	p(\theta \mid S,Y_t) \propto p(S,Y_t\mid\theta)\,p(\theta).
\label{e:bayes}
\end{equation}

Let us look at a real example. \fig{f:toy-example}a shows a histogram of the scores obtained from classifier on a public dataset (see \secn{s:exps} for more information about the datasets we use). At first glance, it is difficult to guess the performance of the classifier unless the oracle provides a lot of labels. However, if we \emph{assume} that the scores follow a two-component mixture model as in \eqref{e:likelihood}, with a Gamma distribution for the $y_i=0$ and a Normal distribution for the $y_i=1$ component, then there is a only a narrow choice of $\theta$ that can explain the scores with high likelihood; the red and green curves in \fig{f:toy-example}a show such a high probability hypothesis. As we will see in the next section, the posterior on $\theta$ can be used to estimate the performance of the classifier.

\begin{figure*}[t]
\begin{center}
\includegraphics[width=\textwidth]{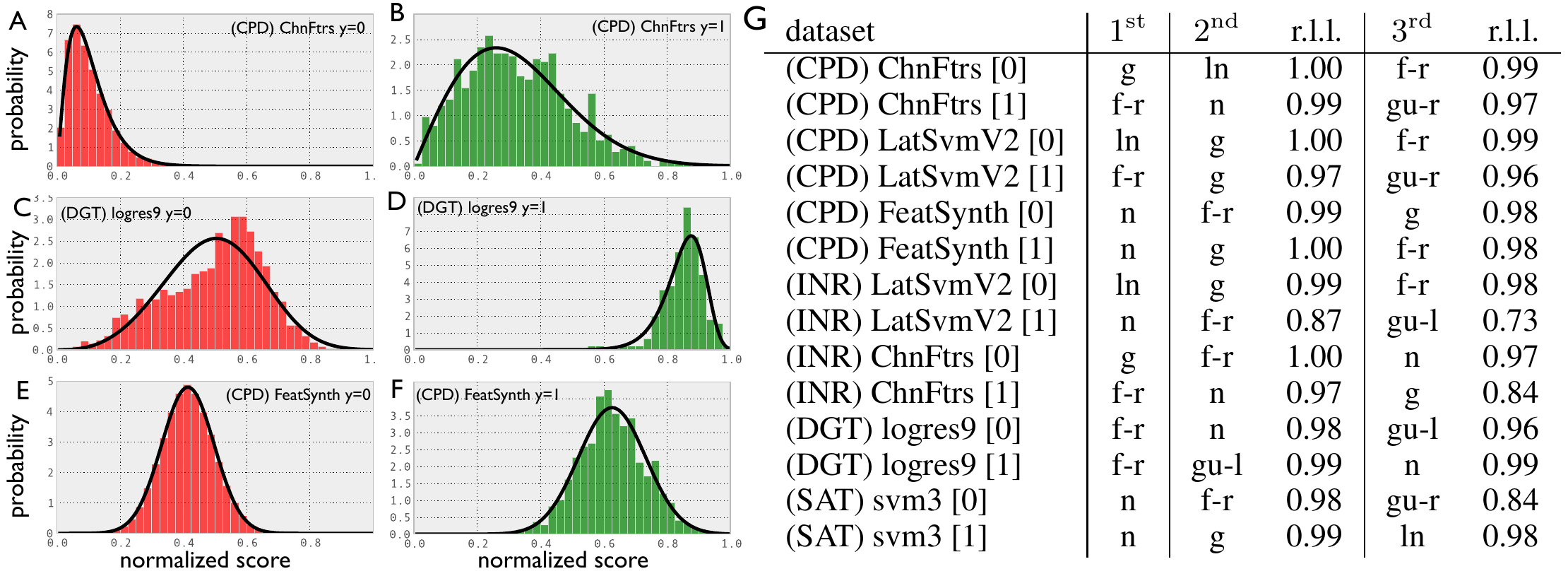}
\caption{Modeling class-conditional score densities by standard parametric distributions. \textbf{A}--\textbf{F}: Standard parametric distributions $p_y(s\mid\theta_y)$ (black solid curve) fitted to the class conditional scores for a few example datasets and classifiers. The score distributions are shown as histograms. In all cases, we normalized the scores to be in the interval $s_i\in(0,1]$, and made the truncation at $s=0$ for the truncated distributions. See \secn{s:fit-dists} for more information. 
\textbf{G}: Comparison of standard parametric distributions best representing empirical class-conditional score distributions (for a subset of the 78 cases we tried). Each row shows the top-3 distributions, i.e. explaining the class-conditional scores with highest likelihood, for different combinations of datasets, classifiers and the class-labels (shown in brackets, $y=0$ or $y=1$). The distribution families we tried included (with abbreviations used in last three columns in parentheses) the truncated Normal (n), truncated Student's t (t), Gamma (g), log-normal (ln), left- and right-skewed Gumbel (g-l and g-r), Gompertz (gz), and Frechet right (f-r) distribution. The last and second to last column show the relative log likelihood (r.l.l.) with respect to the best ($1^{\rm{st}}$) distribution. Two densities, truncated Normal and Gamma, are either top or indistinguishable from the top in all the datasets we tried.}
\label{f:fit-dists}
\end{center}
\end{figure*}

\section{Estimating performance}
Most performance measures can be computed directly from the model parameters $\theta$. For example, two often used performance measures are the precision $P(\tau;\theta)$ and recall $R(\tau;\theta)$ at a particular score threshold $\tau$. We can define these quantities in terms of the conditional distributions $p_y(s_i \mid \theta_y)$. Recall is defined as the fraction of the positive, i.e. $y_i=1$, examples that have scores above a given threshold,
\begin{equation}
	R(\tau;\theta) = \int_\tau^{\infty} p_1(s \mid \theta_1) \, ds.
\end{equation}
Precision is defined to be the fraction of all examples with scores above a given threshold that are positive,
\begin{equation}
	P(\tau;\theta) = \frac{\pi R(\tau;\theta)}
		{\pi R(\tau;\theta) + (1-\pi)\int_\tau^{\infty} p_0(s \mid \theta_0) \, ds}.
\end{equation}
We can also compute the precision at a given level of recall by inverting $R(\tau;\theta)$, i.e. $P_r(r;\theta) = P(R^{-1}(r;\theta);\theta)$ for some recall $r$. Other performance measures, such as the equal error rate, true positive rate, true negative rate, sensitivity, specificity, and the ROC can be computed from $\theta$ in a similar manner.

The posterior on $\theta$ can also be used to obtain confidence bounds on the performance of the classifier. For example, for some choice of parameters $\theta$, the precision and recall can be computed for a range of score thresholds $\tau$ to obtain a curve (see solid curves in \fig{f:toy-example}b). Similarly, given the posterior on $\theta$, the distribution of $P(\tau;\theta)$ and $R(\tau;\theta)$ can be computed for a fixed $\tau$ to obtain confidence intervals (shown as colored bands in \fig{f:toy-example}b). The same reasoning can be applied to the precision-recall curve: for some recall $r$, the distribution of precisions, found using $P_r(r; \theta)$ can be used to compute confidence intervals on the curve (see \fig{f:toy-example}c).

While the approach of estimating performance based purely on the estimate of $\theta$ works well in limit when the number of data items $N\rightarrow \infty$, it has some drawbacks when $N$ is small (on the order of $10^3-10^4$) and $\pi$ is unbalanced, in which case finite-sample effects come into play. This is especially the case when the number of positive examples is very small, say 10--100, in which case the performance curve will be very jagged. Since the previous approach views the scores (and the associated labels) as a finite sample from $p(S,Y\mid\theta)$, there will always be uncertainty in the performance estimate. When all items have been labeled by the oracle, the remaining uncertainty in the performance represents the variability
in sampling $(S,Y)$ from $p(S,Y\mid\theta)$. In practice, however, the question that is often asked is, ``What is our best guess for the classifier performance on this particular test set?'' In other words, we are interested in the sample performance rather than the population performance. Thus, when the oracle has labeled the whole test set, there should not be any uncertainty in the performance; it can simply be computed directly from $(S,Y)$. 

To estimate the sample performance, we need to account for uncertainty in the unlabeled items, $i\in\mathcal{U}_t$. This uncertainty is captured by the distribution of the unobserved labels $Y^{\prime}_t = \{y_i \mid i\in\mathcal{U}_t\}$, found by marginalizing out the model parameters,
\begin{align}
	p(Y^{\prime}_t \mid S, Y_t) 
		&= \int_{\Theta} p(Y^{\prime}_t, \theta \mid S, Y_t)\,d\theta \nonumber \\
		&= \int_{\Theta} p(Y^{\prime}_t \mid \theta) p(\theta \mid S, Y_t)\,d\theta.
\label{e:yu-post}
\end{align}
Here $\Theta$ is the space of all possible parameters. On the second line we rely on the assumption of a mixture model to factor the joint probability distribution on $\theta$ and $Y^{\prime}_t$.

One way to think of this approach is as follows: imagine that we sample $Y^{\prime}_t$ from $p(Y^{\prime}_t \mid S, Y_t)$. We can then use all the labels $Y = Y_t \cup Y^{\prime}_t$ and the scores $S$ to trace out a performance curve (e.g., a precision-recall curve). Now, as we repeat the sampling, each performance curve will look slightly different. Thus, the posterior distribution on $Y^{\prime}_t$ in effect gives us a distribution of performance curves. We can use this distribution to compute quantities such as the expected performance curve, the variance in the curves, and confidence intervals. The main difference between the sample and population performance estimates will be at the tails of the score distribution, $p(S\mid\theta)$, where individual item labels can have a large impact on the performance curve.

\subsection{Sampling from the posterior} \label{s:sampling}
In practice, we cannot compute $p(Y^{\prime}_t \mid S, Y_t)$ in \eqref{e:yu-post} analytically, so we must resort to approximate methods. For some choices of class conditional densities, $p_y(s\mid\theta_0)$, such as when they are Normal distributions, it is possible to carry out the marginalization over $\theta$ in \eqref{e:yu-post} analytically. In that case one could use collapsed Gibbs sampling to sample from the posterior on $Y^{\prime}_t$, as is often done for models involving the Dirichlet process \cite{MacEachern94}. A more generally applicable method, which we will describe here, is to split the sampling into three steps: (a) sample $\bar{\theta}$ from $p(\theta \mid S, Y_t)$, (b) fix the mixture parameters to $\bar{\theta}$ and sample the labels $Y^{\prime}_t$ given their associated scores, and (c) compute the performance, such as precision and recall, for all score thresholds $\tau\in S$. By repeating these three steps, we can obtain a sample from the distribution over the performance curves.

The first step, sampling from the posterior $p(\theta \mid S, Y_t)$, can be carried out using importance sampling (IS). We experimented with Metropolis-Hastings and Hamiltonian Monte Carlo \cite{Neal10}, but we found that IS worked well for this problem, required less parameter tuning, and was much faster. In IS, we sample from a proposal distribution $q(\theta)$ in order to estimate properties of the desired distribution $p(\theta \mid S, Y_t)$. Suppose we draw $M$ samples of $\theta$ from $q(\theta)$ to get $\bar{\Theta} = \{\bar{\theta}^1, \ldots, \bar{\theta}^M\}$. Then, we can approximate expectations of some function $g(\cdot)$ using the weighted function evaluations, i.e. $E[g] \simeq \sum_{m=1}^M w_m g(\bar{\theta}^m)$. The weights $w_m\in W$ correct for the bias introduced by sampling from $q(\theta)$ and are defined as,
\begin{equation}
	w_m = \frac{p(\bar{\theta}^m \mid S, Y_t)/q(\bar{\theta}^m)}{\sum_l p(\bar{\theta}^l \mid S, Y_t)/q(\bar{\theta}^l)}.
\end{equation}

For the datasets in this paper, we found that the state-space around the MAP estimate\footnote{We used BFGS-B \cite{ByrdEtal95} to carry out the optimization. To avoid local maxima, we used multiple starting points.} of $\theta$,
\begin{equation}
	\theta^\star = \arg\max_\theta p(\theta \mid S, Y_l),
\label{e:map}
\end{equation}
was well approximated by a multivariate Normal distribution. Hence, for the proposal distribution we used,
\begin{equation}
	q(\theta) = \mathcal{N}(\theta \mid \mu_q, \Sigma_q).
\end{equation}
To simplify things further, we used a diagonal covariance matrix, $\Sigma_q$. The elements along the diagonal of $\Sigma_q$ were found by fitting a univariate Normal locally to $p(\theta \mid S, Y_t)$ along each dimension of $\theta$ while the other elements were fixed at their MAP-estimates. The mean of the proposal distribution, $\mu_q$, was set to the MAP estimate of $\theta$.

We now have all steps needed to estimate the performance of the classifier, given the scores $S$ and some labels $Y_t$ obtained from the oracle:
\begin{enumerate}
\item Find the MAP estimate $\mu_q$ of $\theta$ using \eqref{e:map}.
\item Fit a proposal distribution $q(\theta)$ to $p(\theta \mid S, Y_t)$ locally around $\mu_q$.
\item Sample $M$ instances of $\theta$, $\bar{\Theta} = \{\bar{\theta}^1, \ldots, \bar{\theta}^M\}$, from $q(\theta)$ and calculate the weights $w_m\in W$.
\item For each $\bar{\theta}^m \in \bar{\Theta}$, sample the labels for $i\in\mathcal{U}_t$ to get $\bar{Y}^{\prime}_t=\{\bar{Y}^{\prime}_{t,1}, \ldots, \bar{Y}^{\prime}_{t,M}\}$.
\item Estimate performance measures using the scores $S$, labels $\bar{Y}_{t,m} = Y_t \cup \bar{Y}^{\prime}_{t,m}$ and weights $w_m\in W$.
\end{enumerate}
\section{Experiments} \label{s:exps}

\begin{figure*}[t]
\begin{center}
\includegraphics[width=\textwidth]{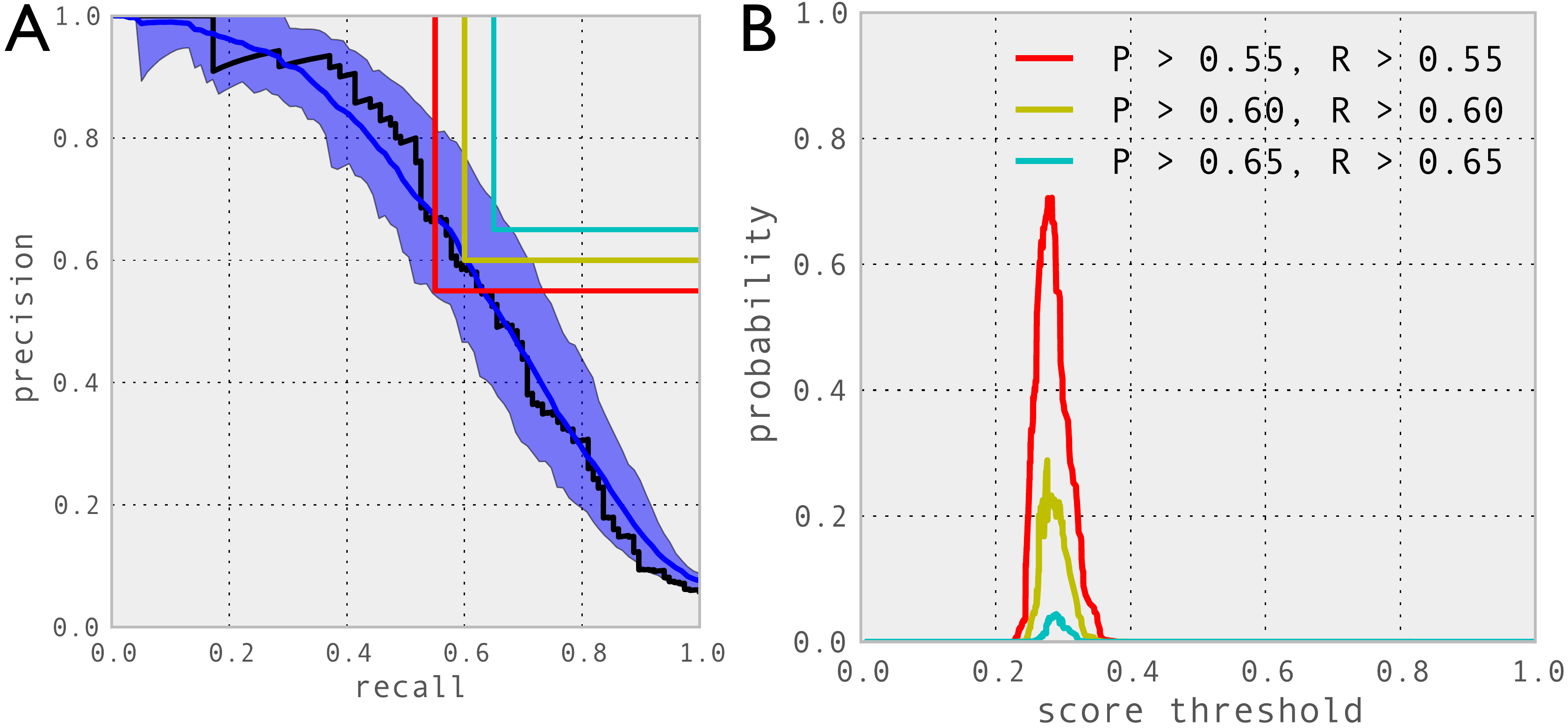}
\caption{Recalibrating the classifier by estimating the probability that a condition is met. \textbf{A}: The conditions in panel B shown as colored ``boxes,'' e.g. the yellow curve shows the condition that the precision $P>0.5$ and recall $R>0.5$. The blue curve and confidence band show SPE applied to the ChnFtrs detector on the CPD dataset with 100 observed labels (black curve is ground truth). \textbf{B}: Probability that the conditions shown in A are satisfied for different score thresholds. Based on a curve like this, a practitioner can ``recalibrate'' a pre-trained classifier by picking a threshold for new dataset such that some pre-defined criteria (e.g. in terms of precision and recall) are met.}
\label{f:thresh}
\end{center}
\end{figure*}

\subsection{Datasets} \label{s:datasets}
We surveyed the literature for published classifier scores with ground truth labels. One such dataset that we found was the Caltech Pedestrian Dataset\footnote{Downloaded from \scriptsize{\url{http://www.vision.caltech.edu/Image_Datasets/CaltechPedestrians/}}.} (CPD), for which both detector scores and ground truth labels are available for a wide variety of detectors \cite{DollarEtal11}. Moreover, the CPD website also has scores and labels available, using the same detectors, for other pedestrian detection datasets, such as the INRIA (abbr. INR) dataset \cite{DalalTriggs05}.

We made use of the detections in the CPD and INR datasets as if they were classifier outputs. To some extent, these detectors are in fact classifiers, in that they use the sliding-window technique for object detection. Here, windows are extracted at different locations and scales in the image, and each window is classified using a pedestrian classifier (with the caveat that there is often some extra post-processing steps carried out, such as non-maximum suppression to reduce the number of false positive detections). For our experiments, we show the results on detectors and datasets to highlight both the advantages and drawbacks with using SPE. To make experiments go faster, we sampled the datasets randomly to have between 800--2,000 items. See \cite{DollarEtal11} for references to all detectors.

To complement the pedestrian datasets, we also used a basic linear SVM classifier and a logistic regression classifier on the ``optdigits'' (abbr. DGT) and ``sat'' (SAT) datasets from the UCI Machine Learning Repository \cite{FrankAsunicon10}. Since both datasets are multiclass, but our method only handles binary classification, we chose one category for $y=1$ and grouped the others into $y=0$. Thus, each multi-class dataset was turned into multiple binary datasets. Planned future work includes extending our approach to multiclass classifiers. In the figures, the naming convention is as follows: ``svm3'' is used to mean that the SVM classifier was used with category 3 in the dataset being assigned to the $y=1$ class, and ``logres9'' denotes that the logistic regression classifier was used with category 9 being the $y=1$ class, and so on. The datasets had 1,800--2,000 items each.


\subsection{Choosing class conditionals} \label{s:fit-dists}
So far we have not discussed in detail which distribution families to use for the class conditional $p_y(s\mid\theta_y)$ distributions. To find out which parametric distributions are appropriate for modeling the score class-conditionals, we took the classifier scores  and split them into two groups, one for $y_i=0$ and one for $y_i=1$. We used MLE to fit different families of probability distributions (see \fig{f:fit-dists}  for a list of distributions) on 80\% of the data (sampled randomly) in each group. We then ranked the distributions by the log likelihood of the remaining 20\% of the data (given the MLE-fitted parameters). In total, we carried out this procedure on 78 class conditionals from the different datasets and classifiers.

\fig{f:fit-dists}G shows the top-3 distributions that explained the class-conditional scores with highest likelihood for a selection of the datasets and classifiers. We found that the truncated Normal distribution was in the top-3 list for 48/78 dataset class-conditionals, and that the Gamma distribution was in the top-3 list 53/78 times; at least one of the two distributions were always in the top-3 list. \fig{f:fit-dists}A--F show some examples of the fitted distributions. In some cases, like \fig{f:fit-dists}C, a mixture model would have provided a better fit than the simple distributions we tried. That said, we found that truncated Normal and Gamma distributions were good choices for most of the datasets. 

Since we use a Bayesian approach in equation \eqref{e:bayes}, we must also define a prior on $\theta$. The prior will vary depending on which distribution is chosen, and it should be chosen based on what we know about the data and the classifier. As an example, for the truncated Normal distribution, we use a Normal and a Gamma distribution as priors on the mean and standard deviation respectively (since we use sampling for inference, we are not limited to conjugate priors). As a prior on the mixture weight $\pi$, we use a Beta distribution.

In some situations when little is known about the classifier, it makes sense to try different kinds of class-conditional distributions. One heuristic, which we found worked well in our experiments, is to try different combinations of distributions for $p_0$ and $p_1$, and then choose the combination achieving the highest maximum likelihood on the labeled and unlabeled data.

\subsection{Applying SPE} \label{s:apply-spe}
\fig{f:q-scheme-qual} shows SPE applied to different datasets. The left-most plots show the estimation error, as measured by the area between the true and predicted precision-recall curves, versus the number of labels sampled. The datasets in \fig{f:q-scheme-qual}A--B and C--D were chosen to highlight the strengths and weaknesses of using SPE. \fig{f:q-scheme-qual}A shows SPE applied to the ChnFtrs detector in the CPD dataset. Already at 20 sampled labels, the estimate is very close (see \fig{f:q-scheme-qual}B). In a few cases, e.g. in \fig{f:q-scheme-qual}C--D (logres8 on the DGT dataset), SPE does not fare as well. While SPE performs as well as the naive method in terms of estimation error, the score distribution is not well explained by the assumptions of the model, so there is a bias in the prediction. That said, despite the fact that SPE is biased in \fig{f:q-scheme-qual}D, it is still far better than the naive method for 100 labels. Ultimately, the accuracy of SPE depends on how well the score data fit the assumptions in \secn{s:model}.

\fig{f:q-scheme-qual}E compares the estimation error of SPE to the naive method for different datasets, when only 20 labels are known. In almost all cases, SPE performs significantly better. Moreover, the variances in the SPE estimates are smaller than those of the naive method.

\subsection{Classifier recalibration} \label{s:recalib}
Applying SPE to a test dataset allows us to ``recalibrate'' the classifier to that dataset. Unlike previous work on classifier calibration \cite{Bennett02,Platt99}, SPE does not require all items to be labeled. For each unlabeled data item, we can compute the probability that it belongs to the $y=1$ class by calculating the empirical expectation from the samples, i.e. $\hat{p}(y_i=1) = E\left[y_i=1 \mid S, Y_t\right]$. 

Similarly, we can choose a threshold $\tau$ to use with the classifier $\bar{h}(x_i; \tau)$ based on some pre-determined criteria. For example, the requirement might be that the classifier performs with recall $R(\tau)>\hat{r}$ and precision $P(\tau)>\hat{p}$. In that case, we define a condition $C(\tau) = \left[R(\tau)>\hat{r}\wedge P(\tau)>\hat{p}\right]$. Then, for each $\tau$, we find the probability that the condition is satisfied by calculating the expectation $\hat{p}(C(\tau)=1)=E\left[C(\tau)\right]$ over the unlabeled items $Y_t^\prime$. \fig{f:thresh} shows the probability that $C(\tau)$ is satisfied at different values of $\tau$. Thus, this approach can be used to choose new thresholds for different datasets.

\section{Related work}
Previous approaches for estimating classifier performance with few labels falls into two categories: stratified sampling and active estimation using importance sampling. Bennett and Carvalho~\cite{BennettCarvalho10} suggested that the accuracy of classifiers can be estimated cost-effectively by dividing the data into disjoint strata based on the item scores, and proposed an online algorithm for sampling from the strata. This work has since been generalized to other classifier performance metrics, such as precision and recall~\cite{DruckMcCallum11}. Sawade et al. proposed instead to use importance sampling to focus labeling effort on data items with high classifier uncertainty, and applied it to standard loss functions~\cite{SawadeEtal10a} and F-measures~\cite{SawadeEtal10b}. While both of these approaches assume that the classifier threshold $\tau$ is fixed (see \secn{s:model}) and that a single scalar performance measure is desired, SPE can be applied to the \emph{tradeoff} between different performance measures in the form of performance curves.

Fitting mixture models to the class-conditional score distributions has been studied in previous work with the goal of obtaining smooth performance curves. Gu et al.~\cite{GuEtal08} and Hellmich et al.~\cite{HellmichEtal99} showed how a two-component Gaussian mixture model can be used to obtain accurate ROC curves in different settings. Erkanli et al.~\cite{ErkanliEtal06} extended this work by fitting mixtures of Dirichlet process priors to the class-conditional distributions. This allowed them to provide smooth performance estimates even when the class-conditional distributions could not be explained by standard parametric distributions. Similarly, previous work on classifier calibration has involved fitting mixture models to score distributions~\cite{Bennett02,Platt99}. In contrast to previous work, which require all data items to be labeled, SPE also makes use of the unlabeled data. This semisupervised approach allows SPE to estimate classifier performance with very few labels, or when the proportions of positive and negative examples are very unbalanced.

\section{Discussion}
We explored the problem of estimating classifier performance from few labeled items. We propose using mixtures of two densities to model the scores of classifiers. This allows us to predict performance curves even when a very small number (none in the limit) of the samples are labeled. Using four public datasets, and multiple classifiers, we showed that classifier score distributions can often be well approximated by two-component mixture models with standard parametric component distributions, such as truncated Normal and Gamma distributions. We demonstrated how our model, Semisupervised Performance Evaluation (SPE), can be used to estimate classifier performance, with confidence intervals, using only a few labeled examples. We presented a sampling scheme based on importance sampling for efficient inference. 

This line of research opens up many interesting avenues for future exploration. For example, is it possible to do unbiased active querying, so that the oracle is asked to label the most informative examples? One possibility in this direction would be to employ importance weighted active sampling techniques~\cite{BeygelzimerEtal08,DasguptaHsu08}, so similar in spirit to \cite{SawadeEtal10a,SawadeEtal10b} but for performance curves. Another future direction would be to extend SPE to multi-component mixture models and multiclass problems. That said, as shown by our experiments, SPE already works well for a broad range of classifiers and datasets, and can estimate classifier performance with as few as 10 labels (see \fig{f:toy-example}).

\bibliography{paper}
\bibliographystyle{plain}

\end{document}